# MDPs with Unawareness

**Joseph Y. Halpern     Nan Rong     Ashutosh Saxena**
Computer Science Department
Cornell University
Ithaca, NY 14853
{halpern | rongnan | asaxena}@cs.cornell.edu

## Abstract

Markov decision processes (MDPs) are widely used for modeling decision-making problems in robotics, automated control, and economics. Traditional MDPs assume that the decision maker (DM) knows all states and actions. However, this may not be true in many situations of interest. We define a new framework, *MDPs with unawareness* (MDPUs) to deal with the possibilities that a DM may not be aware of all possible actions. We provide a complete characterization of when a DM can learn to play near-optimally in an MDPU, and give an algorithm that learns to play near-optimally when it is possible to do so, as efficiently as possible. In particular, we characterize when a near-optimal solution can be found in polynomial time.

## 1  INTRODUCTION

Markov decision processes (MDPs) [2] have been used in a wide variety of settings to model decision making. The description of an MDP includes a set $S$ of possible states and a set $A$ of actions. Unfortunately, in many decision problems of interest, the decision maker (DM) does not know the state space, and is unaware of possible actions she can perform. For example, someone buying insurance may not be aware of all possible contingencies; someone playing a video game may not be aware of all the actions she is allowed to perform nor of all states in the game.

The fact that the DM may not be aware of all states does not cause major problems. If an action leads to a new state and the set of possible actions is known, we can use standard techniques (discussed below) to decide what to do next. The more interesting issue comes in dealing with actions that the DM may not be aware of. If the DM is not aware of her lack of awareness then it is clear how to proceed—we can simply ignore these actions; they are not on the DM's radar screen. We are interested in a situation where the DM realizes that there are actions (and states) that she is not aware of, and thus will want to explore the MDP. We model this by using a special *explore* action. As a result of playing this action, the DM might become aware of more actions, whose effect she can then try to understand.

We have been deliberately vague about what it means for a DM to be "unaware" on an action. We have in mind a setting where there is a (possibly large) space $A^*$ of *potential actions*. For example, in a video game, the space of potential actions may consist of all possible inputs from all input devices combined (e.g., all combinations of mouse movements, presses of keys on the keyboard, and eye movements in front of the webcam); if a DM is trying to prove a theorem, at least in principle, all possible proof techniques can be described in English, so the space of potential actions can be viewed as a subset of the set of English texts. The space $A$ of *actual actions* is the (typically small) subset of $A^*$ that are the "useful actions". For example, in a video game, these would be the combinations of arrow presses (and perhaps head movements) that have an appreciable effect on the game. Of course, $A^*$ may not describe how the DM conceives of the potential acts. For example, a first-time video-game player may consider the action space to include only presses of the arrow keys, and be completely unaware that eye movement is an action. Similarly, a mathematician trying to find a proof probably does not think of herself as searching in a space of English texts; she is more likely to be exploring the space of "proof techniques". A sophisticated mathematician or video game player will have a better understanding of the space that she views herself as exploring. Moreover, the space of potential actions may change over time, as the DM becomes more sophisticated. Thus, we do not explicitly describe $A^*$ in our formal model, and abstract the process of exploration by just having an *explore* action.

This type of exploration occurs all the time. In video games, first-time players often try to learn the game by exploring the space of moves, without reading the instructions (and thus, without being aware of all the moves they can make). Indeed, in many games, there may not be instructions at all (even though players can often learn what moves are available by checking various sites on the web). Mathematicians trying to generate new approaches to proving a theorem can be viewed as exploring the space of proof techniques. More practically, in robotics, if we take an action to be a "useful" sequence of basic moves, the space of potential actions is often huge. For instance, most humanoid robots (such as Honda Asimo robot [13]) have more than 20 degrees of freedom; in such a large space, while robot designers can hand-program a few basic actions (e.g., as walking on a level surface), it is practically impossible to do so for other general scenarios (e.g., walking on uneven rocks). Conceptually, it is useful to think of the designer as not being aware of the actions that can be performed. Exploration is almost surely necessary to discover new actions necessary to enable the robot to perform the new tasks.

Given the prevalence of MPDUs—*MDPs with unawareness*, the problem of learning to play well in an MDPU becomes of interest. There has already been a great deal of work on learning to play optimally in an MDP. Kearns and Singh [11] gave an algorithm called $E^3$ that converges to near-optimal play in polynomial time. Brafman and Tennenholtz [3] later gave an elegant algorithm they called RMAX that converges to near-optimal play in polynomial time not just in MDPs, but in a number of adversarial settings. Can we learn to play near-optimally in an MDPU? (By "near-optimal play", we mean near-optimal play in the actual MDP.) In the earlier work, near-optimal play involved learning the effects of actions (that is, the transition probabilities induced by the action). In our setting, the DM still has to learn the transition probabilities, but also has to learn what actions are available.

Perhaps not surprisingly, we show that how effectively the DM can learn optimal play in an MDPU depends on the probability of discovering new actions. For example, if it is too low, then we can never learn to play near-optimally. If it is a little higher, then the DM can learn to play near-optimally, but it may take exponential time. If it is sufficiently high, then the DM can learn to play near-optimally in polynomial time. We give an expression whose value, under minimal assumptions, completely characterizes when the DM can learn to play optimally, and how long it will take. Moreover, we show that a modification of the RMAX algorithm (that we call URMAX) can learn to play near-optimally if it is possible to do so.

There is a subtlety here. Not only might the DM not be aware of what actions can be performed in a given state, she may be unaware of *how many* actions can be performed. Thus, for example, in a state where she has discovered five actions, she may not know whether she has discovered all the actions (in which case she should not explore further) or there are more actions to be found (in which case she should). In cases where the DM knows that there is only one action to be discovered, and what its payoff is, it is still possible that the DM never learns to play optimally. Our impossibility results and lower bound hold even in this case. (For example, if the action to be discovered is a proof that P ≠ NP, the DM may know that the action has a high payoff; she just does not know what that action is.) On the other hand, URMAX works even if the DM does not know how many actions there are to be discovered.

There has been a great deal of recent work on awareness in the game theory literature (see, for example, [5, 8, 10]). There has also been work on MDPs with a large action space (see, for example [4, 9]), and on finding new actions once exploration is initiated [1]. None of these papers, however, considers the problem of learning in the presence of lack of awareness.

The rest of the paper is organized as follows. In Section 2, we review the work on learning to play optimally in MDPs. In Section 3, we describe our model of MDPUs. We give our impossibility results and lower bounds in Section 4. In Section 5, we present a general learning algorithm (adapted from R-MAX) for MDPU problems, and give upper bounds. We conclude in Section 6. Missing proofs can be found in the full paper.

## 2 PRELIMINARIES

**MDPs:** An MDP is a tuple $M = (S, A, P, R)$, where $S$ is a finite set of states; $A$ is a finite set of actions; $P : (S \times S \times A) \to [0, 1]$ is the transition probability function, where $P(s, s', a)$ gives the transition probability from state $s$ to state $s'$ with action $a$; and $R : (S \times S \times A) \to \mathbb{R}^+$ is the reward function, where $R(s, s', a)$ gives the reward for playing action $a$ at state $s$ and transiting to state $s'$. Since $P$ is a probability function, we have $\sum_{s' \in S} P(s, s', a) = 1$ for all $s \in S$ and $a \in A$. A *policy* in an MDP $(S, A, P, R)$ is a function from histories to actions in $A$. Given an MDP $M = (S, A, P, R)$, let $U_M(s, \pi, T)$ denote the expected $T$-step undiscounted average reward of policy $\pi$ started in state $s$—that is, the expected total reward of running $\pi$ for $T$ steps, divided by $T$. Let $U_M(s, \pi) = \lim_{T \to \infty} U_M(s, \pi, T)$, and let $U_M(\pi) = \min_{s \in S} U_M(s, \pi)$.

**The mixing time:** For a policy $\pi$ such that $U_M(\pi) = \alpha$, it may take a long time for $\pi$ to get an expected payoff of $\alpha$. For example, if getting a high reward involves reaching a particular state $s^*$, and the probability of reaching $s^*$ from some state $s$ is low, then the time to get the high reward will be high. To deal with this, Kearns and Singh [11] argue that the running time of a learning algorithm should be compared to the time that an algorithm with full information takes to get a comparable reward. Define the $\epsilon$-*return mixing time of policy* $\pi$ to be the smallest value of $T$ such that $\pi$ guarantees an expected payoff of at least $U(\pi) - \epsilon$;

that is, it is the least $T$ such that $U(s, \pi, t) \geq U(\pi) - \epsilon$ for all states $s$ and times $t \geq T$. Let $\Pi(\epsilon, T)$ consist of all policies whose $\epsilon$-mixing time is at most $T$. Let $\text{Opt}(M, \epsilon, T) = \max_{\pi \in \Pi(\epsilon, T)} U_M(\pi)$.

RMAX: We now briefly describe the RMAX algorithm [3]. RMAX assumes that the DM knows all the actions that can be played in the game, but needs to learn the transition probabilities and reward function associated with each action. It does not assume that the DM knows all states; new states might be discovered when playing actions at known states. RMAX follows an implicit "explore or exploit" mechanism that is biased towards exploration. Here is the RMAX algorithm:

RMAX($|S|, |A|, R_{\max}, T, \epsilon, \delta, s_0$):
Set $K_1(T) := \max((\lceil \frac{4|S|T R_{\max}}{\epsilon} \rceil)^3, \lceil -6 \ln^3(\frac{\delta}{6|S||A|^2}) \rceil) + 1$
Set $M' := M^0$ (the initial approximation described below)
Compute an optimal policy $\pi'$ for $M'$
Repeat until all action/state pairs $(s, a)$ are *known*
    Play $\pi'$ starting in state $s_0$ for $T$ steps or until some new state-action pair $(s, a)$ is known
    **if** $(s, a)$ has just become *known* **then** update $M'$ so that the transition probabilities for $(s, a)$ are the observed frequencies and the rewards for playing $(s, a)$ are those that have been observed.
    Compute the optimal policy $\pi'$ for $M'$
Return $\pi'$.

Here $R_{\max}$ is the maximum possible reward; $\epsilon > 0$; $0 < \delta < 1$; $T$ is the $\epsilon$-return mixing time; $K_1(T)$ represents the number of visits required to approximate a transition function; a state-action pair $(s, a)$ is said to be *known* only if it has been played $K_1(T)$ times. RMAX proceeds in iterations, and $M'$ is the current "approximation" to the true MDP. $M'$ consists state set $S$ and a dummy state $s_d$. The transition and reward functions in $M'$ may be different from those of the actual MDP. In the initial approximation $M^0$, the transition and reward functions are trivial: when an action $a$ is taken in any state $s$ (including the dummy state $s_d$), with probability 1 there is a transition to $s_d$, with reward $R_{\max}$.

Brafman and Tennenholtz [3] show that RMAX($|S|, |A|, R_{\max}, T, \epsilon, \delta, s_0$) learns a policy with expected payoff within $\epsilon$ of $\text{Opt}(M, \epsilon, T)$ with probability greater than $1 - \delta$, no matter what state $s_0$ it starts in, in time polynomial in $|S|, |A|, T, 1/\delta$, and $1/\epsilon$. What makes RMAX work is that in each iteration, it either achieves a near-optimal reward with respect to the real model or learns an unknown transition with high probability. Since there are only polynomially-many $(s, a)$ pairs (in the number of states and actions) to learn, and each transition entry requires $K_1(T)$ samples, where $K_1(T)$ is polynomial in the number of states and actions, $1/\epsilon$, $1/\delta$, and the $\epsilon$-return mixing time $T$, RMAX clearly runs in time polynomial in these parameters. In the case that the $\epsilon$-return mixing time $T$ is not known, RMAX starts with $T = 1$, then considers $T = 2, T = 3$, and so on.

## 3 MDPS WITH UNAWARENESS

Intuitively, an MDPU is like a standard MDP except that the player is initially aware of only a subset of the complete set of states and actions. To reflect the fact that new states and actions may be learned during the game, the model provides a special *explore* action. By playing this action, the DM may become aware of actions that she was previously unaware of. The model includes a *discovery probability function* characterizing the likelihood that a new action will be discovered. At any moment in game, the DM can perform only actions that she is currently aware of.

**Definition 3.1 :** An MDPU is a tuple $M = (S, A, S_0, a_0, g_A, g_0, P, D, R, R^+, R^-)$, where

- $S$, the set of states in the underlying MDP;
- $A$, the set of actions in the underlying MDP;
- $S_0 \subseteq S$ is the set of states that the DM is initially aware of;
- $a_0 \notin A$ is the *explore* action;
- $g_A : S \to 2^A$, where $g_A(s)$ is the set of actions that can be performed at $s$ other than $a_0$ ($a_0$ can be performed in every state);
- $g_0 : S_0 \to 2^A$, where $g_0(s) \subseteq g_A(s)$ is the set of actions that the DM is aware of at state $s$ (the DM is always aware of $a_0$);
- $P : \cup_{s \in S}(\{s\} \times S \times g_A(s)) \to [0, 1]$ is the transition probability function (as usual, we require that $\sum_{s' \in S} P(s, s', a) = 1$ if $a \in g_A(s)$);
- $D : \mathbb{N} \times \mathbb{N} \times S \to [0, 1]$ is the discovery probability function. $D(j, t, s)$ gives the probability of discovering a new action in state $s \in S$ given that there are $j$ actions to be discovered and $a_0$ has already been played $t - 1$ times in $s$ without a new action being discovered (see below for further discussion);
- $R : \cup_{s \in S}(\{s\} \times S \times g_A(s)) \to \mathbb{R}^+$ is the reward function;[1]
- $R^+ : S \to \mathbb{R}^+$ and $R^- : S \to \mathbb{R}^+$ give the exploration reward for playing $a_0$ at state $s \in S$ and discovering (resp., not discovering) a new action (see below for further discussion).

Given $S_0$ and $g_0$, we abuse notation and take $A_0 = \cup_{s \in s_0} g_0(s)$; that is, $A_0$ is the set of actions that the DM is aware of.

---

[1] We assume without loss of generality that all payoffs are non-negative. If not, we can shift all rewards by a positive value so that all payoffs become non-negative.

Just like a standard MDP, an MDPU has a state space $S$, action space $A$, transition probability function $P$, and reward function $R$.[2] Note that we do not give the transition function for the explore action $a_0$ above; since we assume that $a_0$ does not result in a state change (although new actions might be discovered when $a_0$ is played), for each state $s \in S$, we have $P(s, s, a_0) = 1$. The new features here involve dealing with $a_0$. We need to quantify how hard it is to discover a new action. Intuitively, this should in general depend on how many actions there are to be discovered, and how long the DM has been trying to find a new action. For example, if the DM has in fact found all the actions, then this probability is clearly 0. Since the DM is not assumed to know in general how many actions there are to be found, all we can do is give what we view as the DM's subjective probability of finding a new action, given that there are $j$ actions to be found. Note that even if the DM does not know the number of actions, she can still condition on there being $j$ actions. In general, we also expect this probability to depend on how long the DM has been trying to find a new action. This probability is captured by $D(j, t, s)$.

We assume that $D(j, t, s)$ is nondecreasing as a function of $j$: with more actions available, it is easier to find a new one. How $D(j, t, s)$ varies with $t$ depends on the problem. For example, if the DM is searching for the on/off button on her new iPhone which is guaranteed to be found in a limited surface area, then $D(j, t, s)$ should increase as a function of $t$. The more possibilities have been eliminated, the more likely it is that the DM will find the button when the next possibility is tested. On the other hand, if the DM is searching for a proof, then the longer she searches without finding one, the more discouraged she will get; she will believe that it is more likely that no proof exists. In this case, we would expect $D(j, t, s)$ to decrease as a function of $t$. Finally, if we think of the *explore* action as doing a random test in some space of potential actions, the probability of finding a new action is a constant, independent of $t$. In the sequel, we assume for ease of exposition that $D(j, t, s)$ is independent of $s$, so we write $D(j, t)$ rather than $D(j, t, s)$.

$R^+$ and $R^-$ are the analogues of the reward function $R$ for the *explore* action $a_0$. For example, in a chess game, the *explore* action corresponds to thinking. There is clearly a negative reward to thinking and not discovering a new action—valuable time is lost; we capture this by $R^-(s)$. On the other hand, a player often gets a thrill if a useful action is discovered; and this is captured by $R^+(s)$. It seems reasonable to require that $R^-(s) \leq R^+(s)$, which we do from here on.

When an MDPU starts, $S_0$ represents the set of states that the DM is initially aware of, and $g_0(s)$ represents the set of actions that she is aware of at state $s$. The DM may discover new states when trying out known actions, she may also discover new actions as the explore action $a_0$ is played. At any time, the DM has a current set of states and actions that she is aware of; she can play only actions from the set that she is currently aware of.

In stating our results, we need to be clear about what the inputs to an algorithm for near-optimal play are. We assume that $S_0$, $g_0$, $D$, $R^+$, and $R^-$ are always part of the input to the algorithm. The reward function $R$ is not given, but is part of what is learned. (We could equally well assume that $R$ is given for the actions and states that the DM is aware of; this assumption would have no impact on our results.) Brafman and Tennenholtz [3] assume that the DM is given a bound on the maximum reward, but later show that this information is not needed to learn to play near-optimally in their setting. Our algorithm URMAX does not need to be given a bound on the reward either. Perhaps the most interesting question is what the DM knows about $A$ and $S$. Our lower bounds and impossibility result hold even if the DM knows $|S|$ and $|g_A(s)|$ for all $s \in S$. On the other hand, URMAX requires neither $|S|$ nor $|g_A(s)|$ for $s \in S$. That is, when something cannot be done, knowing the size of the set of states and actions does not help; but when something can be done, it can be done without knowing the size of the set of states and actions.

Formally, we can view the DM's knowledge as the input to the learning algorithm. An MDP $M$ is *compatible with the DM's knowledge* if all the parameters of of $M$ agree with the corresponding parameters that the DM knows about. If the DM knows only $S_0$, $g_0$, $D$, $R^+$, and $R^-$ (we assume that the DM always knows at least this), then every MDP $(S', A', g', P', R')$ where $S_0 \subseteq S'$ and $g_0(s) \subseteq A'(s)$ is compatible with the DM's knowledge. If the DM also knows $|S|$, then we must have $|S'| = |S|$; if the DM knows that $S = S_0$, then we must have $S' = S_0$. We use $R_{\max}$ to denote the maximum possible reward. Thus, if the DM knows $R_{\max}$, then in a compatible MDP, we have $R(s, s', a') \leq R_{\max}$, with equality holding for some transition. (The DM may just know a bound on $R_{\max}$, or not know $R_{\max}$ at all.) If the DM knows $R_{\max}$, we assume that $R^+(s) < R_{\max}$ for all $s \in S$ (for otherwise, the optimal policy for the MDPU becomes trivial: the DM should just get to state $s$ and keep exploring). Brafman and Tennenholtz essentially assume that the DM knows $|A|$, $|S|$, and $R_{\max}$. They say that they believe that the assumption that the DM knows $R_{\max}$ can be removed. It follows from our results that the DM does not need to know any of $|A|$, $|S|$, or $R_{\max}$.

Our theorems talk about whether there is an algorithm for a DM to learn to play near-optimally given some knowledge. We define "near-optimal play" by extending the definitions of [3, 11] to deal with unawareness. In an MDPU, a policy is again a function from histories to actions, but now the action must be one that the DM is aware of at the last state in the history. The DM *can learn to play near-optimally given a state space $S_0$ and some other knowledge* if, for all $\epsilon > 0$, $\delta > 0$, $T$, and $s \in S_0$, the DM can learn a policy $\pi_{\epsilon,\delta,T,s}$ such that, for all MDPs $M$ compatible with the DM's knowledge, there exists a time $t_{M,\epsilon,\delta,T}$ such that, with probability at least $1 - \delta$, $U_M(s, \pi_{\epsilon,\delta,T,s}, t) \geq \text{Opt}(M, \epsilon, T) - \epsilon$ for all $t \geq t_{M,\epsilon,\delta,T}$.[3] The DM *can learn to play near-optimally given some knowledge in polynomial (resp., exponential) time* if, there exists a polynomial (resp., exponential) function $f$ of five arguments such that we can take $t_{M,\epsilon,\delta,T} = f(T, |S|, |A|, 1/\epsilon, 1/\delta)$.

---

[2] It is often assumed that the same actions can be performed in all states. Here we allow slightly more generality by assuming that the actions that can be performed is state-dependent, where the dependence is given by $g$.

## 4 IMPOSSIBILITY RESULTS AND LOWER BOUNDS

The ability to estimate in which cases the DM can learn to play optimally is crucial in many situations. For example, in robotics, if the probability of discovering new actions is so low that it would would require an exponential time to learn to play near-optimally, then the designer of the robot must have human engineers design the actions and not rely on automatic discovery. We begin by trying to understand when it is feasible to learn to play optimally, and then consider how to do so.

We first show that, for some problems, there are no algorithms that can guarantee near-optimal play; in other cases, there are algorithms that will learn to play near-optimally, but will require at least exponential time to do so. These results hold even for problems where the DM knows that there are two actions, already knows one of them, and knows the reward of the other.

**Example 4.1:** Suppose that the DM knows that $S = S_0 = \{s_1\}$, $g_0(s_1) = \{a_1\}$, $|A| = 2$, $P(s_1, s_1, a) = 1$ for all action $a \in A$, $R(s_1, s_1, a_1) = r_1$, $R^+(s_1) = R^-(s_1) = 0$, $D(j, t) = \frac{1}{(t+1)^2}$, and the reward for the optimal policy in the true MDP is $r_2$, where $r_2 > r_1$. Since the DM knows that there is only one state and two actions, the DM knows that in the true MDP, there is an action $a_2$ that she is not aware of such that $R(s_1, s_1, a_2) = r_2$. That is, she knows everything about the true MDP but the action $a_2$. We now show that, given this knowledge, the DM cannot learn to play optimally.

Clearly in the true MDP the optimal policy is to always play $a_2$. However, to play $a_2$, the DM must learn about $a_2$. As we now show, no algorithm can learn about $a_2$ with probability greater than $1/2$, and thus no algorithm can attain an expected return $\geq (r_1 + r_2)/2 = r_2 - (r_2 - r_1)/2$.

Let $E_{t,s}$ denote the event of playing $a_0$ $t$ times at state $s$

---

[3]Note that we allow the policy to depend on the state. However, it must have an expected payoff that is close to that obtained by $M$ no matter what state $M$ is started in.

without discovering a new action, conditional on there being at least one undiscovered action. Since there is exactly one unknown action, and the DM knows this, we have

$$\begin{aligned} Pr(E_{t,s_1}) &= \prod_{t'=1}^{t}(1 - D(1,t')) \\ &= \prod_{t'=1}^{t}\left(1 - \frac{1}{(t'+1)^2}\right) \\ &= \frac{t+2}{2(t+1)} \\ &> \frac{1}{2}. \end{aligned}$$

For the third equality, note that $1 - \frac{1}{(t'+1)^2} = (1 - \frac{1}{t'+1}) \times (1 + \frac{1}{t'+1})$; it follows that $\prod_{t'=1}^{t}\left(1 - \frac{1}{(t'+1)^2}\right) = \left(\frac{1}{2} \times \frac{3}{2}\right) \times \left(\frac{2}{3} \times \frac{4}{3}\right) \times \cdots \times \left(\frac{t}{t+1} \times \frac{t+2}{t+1}\right)$. All terms but the first and last cancel out. Thus, the product is $\frac{t+2}{2(t+1)}$. The inequality above shows that $Pr(E_t, s_1)$ is always strictly greater than 1/2, independent of $t$. In other words, the DM cannot discover the better action $a_2$ with probability greater than $1/2$ no matter how many times $a_0$ is played. It easily follows that the expected reward of any policy is at most $(r_1 + r_2)/2$. Thus, there is no algorithm that learns to play near-optimally. ∎

The problem in Example 4.1 is that the discovery probability is so low that there is a probability bounded away from 0 that some action will not be discovered, no matter how many times $a_0$ is played. The following theorem generalizes Example 4.1, giving a sufficient condition on the failure probability (which we later show is also necessary) that captures the precise sense in which the discovery probability is too low. Intuitively, the theorem says that if the DM is unaware of some acts that can improve her expected reward, and the discovery probability is sufficiently low, where "sufficiently low" means $D(1, t) < 1$ for all $t$ and $\sum_{t=1}^{\infty} D(1, t) < \infty$, then the DM cannot learn to play near-optimally. To make the theorem as strong as possible, we show that the lower bound holds even if the DM has quite a bit of extra information, as characterized in the following definition.

**Definition 4.2:** Define a DM to be *quite knowledgeable* if (in addition to $S_0$, $g_0$, $D$, $R^+$, and $R^-$) she knows $S = S_0$, $|A|$, the transition function $P_0$, the reward function $R_0$ for states in $S_0$ and actions in $A_0$, and $R_{\max}$.

We can now state our theorem. It turns out that there are slightly different conditions on the lower bound depending on whether $|S_0| \geq 2$ or $|S_0| = 1$.

**Theorem 4.3:** *If $D(1, t) < 1$ for all $t$ and $\sum_{t=1}^{\infty} D(1, t) < \infty$, then there exists a constant $c$ such that no algorithm can obtain within $c$ of the optimal reward for all MDPs that are compatible with what the DM knows, even if the DM is quite knowledgeable, provided that $|S_0| \geq 2$, $|A| > |A_0|$, and $R_{\max}$ is greater than the reward of the optimal policy* in the MDP $(S_0, A_0, P_0, R_0)$. If $|S_0| = 1$, the same result holds if $\sum_{t=1}^{\infty} D(j,t) < \infty$, where $j = |A| - |A_0|$.

**Proof:** We construct an MDP $M'' = (S, A'', g'', P'', R'')$ that is compatible with what the DM knows, such that no algorithm can obtain within a constant $c$ of the optimal reward in $M''$. The construction is similar in spirit to that of Example 4.1. Since $|S| \geq 2$, let $s_1$ be a state in $S$. Let $j = |A| - |A_0|$, let $A'' = A_0 \cup \{a_1, \ldots, a_j\}$, where $a_1, \ldots, a_j$ are fresh actions not in $A_0$, let $g''$ be such that $g''(s_1) = g_0(s_1) \cup \{a_1\}$, $g''(s) = A''$, for $s \neq s_1$. That is, there is only one action that the DM is not aware of in state $s_1$, while in all other states, she is unaware of all actions in $A - A_0$. Let $P''(s_1, s_1, a_1) = P''(s, s_1, a) = 1$ for all $a \in A'' - A_0$ and $s \in S$ (note that $P''$ is determined by $P_0$ in all other cases). It is easy to check that $M''$ is compatible with what the DM knows, even if the DM knows that $S = S_0$, knows $|A|$, and knows $R_{\max}$. Let $R''(s_1, s_1, a_1) = R''(s, s_1, a) = R_{\max}$ for all $s \neq s_1$ and $a \in A - A_0$ ($R''$ is determined by $R_0$ in all other cases). By assumption, the reward of the optimal policy in $(S_0, A_0, g_0, P_0, R_0)$ is less than $R_{\max}$, so the optimal policy is clearly to get to state $s_1$ and then to play $a_1$ (giving an average reward of $R_{\max}$ per time unit). Of course, doing this requires learning $a_1$.

As in Example 4.1, we first prove that for $M''$ there exists a constant $d > 0$ such that, with probability $d$, no algorithm will discover action $a_1$ in state $s_1$. The result then follows as in Example 4.1. We leave details to the full paper. ∎

Note that Example 4.1 is a special case of Theorem 4.3, since $\sum_{t=1}^{\infty} \frac{1}{(t+1)^2} < \int_{t=1}^{\infty} \frac{1}{t^2} dt = 1$.

In the next section, we show that if $\sum_{t=1}^{\infty} D(1,t) = \infty$, then there is an algorithm that learns near-optimal play (although the algorithm may not be efficient). Thus, $\sum_{t=1}^{\infty} D(1,t)$ determines whether or not there is an algorithm that learns near-optimal play. We can say even more. If $\sum_{t=1}^{\infty} D(1,t) = \infty$, then the efficiency of the best algorithm for determining near-optimal play depends on how quickly $\sum_{t=1}^{\infty} D(1,t)$ diverges. Specifically, the following theorem shows that if $\sum_{t=1}^{T} D(1,t) \leq f(T)$, where $f : [1, \infty] \to \mathbb{R}$ is an increasing function whose co-domain includes $(0, \infty]$ (so that $f^{-1}(t)$ is well defined for $t \in (0, \infty]$) and $D(1,t) \leq c < 1$ for all $t$, then the DM cannot learn to play near-optimally with probability $\geq 1 - \delta$ in time less than $f^{-1}(c \ln(\delta)/\ln(1-c))$. It follows, for example, that if $f(T) = m_1 \log(T) + m_2$, then it requires time polynomial in $1/\delta$ to learn to play near-optimally with probability greater than $1 - \delta$. For if $f(T) = m_1 \log(T) + m_2$, then $f^{-1}(t) = e^{(t-m_2)/m_1}$, so $f^{-1}(c \ln(\delta)/\ln(1-c)) = f^{-1}(c \ln(1/\delta)/\ln(1/(1-c)))$ has the form $a(1/\delta)^b$ for constants $a, b > 0$. A similar argument shows that if $f(T) = m_1 \ln(\ln(T) + 1) + m_2$, then $f^{-1}(c \ln(1/\delta)/\ln(1/(1-c)))$ has the form $ae^{(1/\delta)^b}$ for constants $a, b > 0$; that is, the running time is exponential in $1/\delta$.

**Theorem 4.4 :** *If $|S_0| \geq 2$, $|A| > |A_0|$, $R_{\max}$ is greater than the reward of the optimal policy in the MDP $(S_0, A_0, P_0, R_0)$, $\sum_{t=1}^{\infty} D(1,t) = \infty$, and there exists a constant $c < 1$ such that $D(1,t) \leq c$ for all $t$, and an increasing function $f : [1, \infty] \to \mathbb{R}$ such that the co-domain of $f$ includes $(0, \infty]$ and $\sum_{t=1}^{T} D(1,t) \leq f(T)$, then for all $\delta$ with $0 < \delta < 1$, there exists a constant $d > 0$ such that no algorithm that runs in time less than $f^{-1}(c \ln(\delta)/\ln(1-c))$ can obtain within $d$ of the optimal reward for all MDPs that are compatible with what the DM knows with probability $\geq 1 - \delta$, even if the DM is quite knowledgeable. If $|S_0| = 1$, the same result holds if $\sum_{t=1}^{T} D(j,t) \leq f(T)$, where $j = |A| - |A_0|$.*

In the next section, we prove that the lower bound of Theorem 4.4 is tight.

## 5 LEARNING TO PLAY NEAR-OPTIMALLY

In this section, we show that a DM can learn to play near-optimally in an MDPU where $\sum_{t=1}^{\infty} D(1,t) = \infty$. Moreover, we show that when $\sum_{t=1}^{\infty} D(1,t) = \infty$, the speed at which $D(1,t)$ decreases determines how quickly the DM can learn to play near-optimally. While the condition $\sum_{t=1}^{\infty} D(1,t) = \infty$ may seem rather special, in fact it arises in many applications of interest. For example, when learning to fly a helicopter [1, 14], the space of potential actions in which the exploration takes place, while four dimensional (resulting from the six degree of freedom of the helicopter), can be discretized and taken to be finite. Thus, if we explore by examining the potential actions uniformly at random, then $D(1,t)$ is constant for all $t$, and hence $\sum_{t=1}^{\infty} D(1,t) = \infty$. Indeed, in this case $\sum_{t=1}^{T} D(1,t)$ is $O(T)$, so it follows from Corollary 5.4 below that we can learn to fly the helicopter near-optimally in polynomial time. The same is true in any situation where the space of potential actions in which the exploration takes place is finite and understood.

We assume throughout this section that $\sum_{t=1}^{\infty} D(1,t) = \infty$. We would like to use an RMAX-like algorithm to learn to play near-optimally in our setting too, but there are two major problems in doing so. The first is that we do not want to assume that the DM knows $|S|$, $|A|$, or $R_{\max}$. We deal with the fact that $|S|$ and $|A|$ are unknown by using essentially the same idea as Kearns and Singh use for dealing with the fact that the true $\epsilon$-mixing time $T$ is unknown: we start with an estimate of the value of $|S|$ and $|A|$, and keep increasing the estimate. Eventually, we get to the right values, and we can compensate for the fact that the payoff return may have been too low up to that point by playing the policy sufficiently often. The idea for dealing with the fact that $R_{\max}$ is not known is similar. We start with an estimate of the value of $R_{\max}$, and recompute the value of $K_1(T)$ and the approximating MDP every time we discover a transition with a reward higher than the current estimate. (We remark that this idea can be applied to RMAX as well.) The second problem is more serious: we need to deal with the fact that not all actions are known, and that we have a special *explore* action. Specifically, we need to come up with an analogue of $K_1(T)$ that describes how many times we should play the explore action $a_0$ in a state $s$, with a goal of discovering all the actions in $s$.

We now describe the URMAX algorithm under the assumption that the DM knows $N$, an upper bound on the state space $S$, $k$, an upper bound on the size of the action space $A$, $R_{\max}$, an upper bound on the true maximum reward, and $T$, an upper bound on the $\epsilon$-return mixing time. To emphasize the dependence on these parameters, we denote the algorithm URMAX($S_0, g_0, D, N, k, R_{\max}, T, \epsilon, \delta, s_0$). (The DM may also know $R^+$ and $R^-$, but the algorithm does not need these inputs.) We later show how to define URMAX($S_0, g_0, D, \epsilon, \delta, s_0$), dropping the assumption that the DM knows $N$, $k$, $T$ and $R_{\max}$.

Define

- $K_1(T) = \max((\lceil \frac{4NTR_{\max}}{\epsilon} \rceil)^3, \lceil 8\ln^3(\frac{8Nk}{\delta}) \rceil) + 1$;
- $K_0 = \min_M \{M : \sum_{t=1}^{M} D(1,t) \geq \ln(4N/\delta)\}$. (Such a $K_0$ always exists if $\sum_{t=1}^{M} D(1,t) = \infty$.)

Just as with RMAX, $K_1(T)$ is a bound on how long the DM needs to get a good estimate of the transition probabilities at each state $s$. Our definition of $K_1(T)$ differs slightly from that of Brafman and Tennenholtz (we have a coefficient 8 rather than 6; the difference turn out to be needed to allow for the fact that we do not know all the actions). As we show below (Lemma 5.1), $K_0$ is a good estimate on how often the *explore* action needs to be played in order to ensure that, with high probability (greater than $1 - \delta/4N$), at least one new action is discovered at a state, if there is a new action to be discovered. Just as with RMAX, we take a pair $(s,a)$ for $a \neq a_0$ to be *known* if it is played $K_1$ times; we take a pair $(s, a_0)$ to be *known* if it is played $K_0$ times.

URMAX($S_0, g_0, D, N, k, R_{\max}, T, \epsilon, \delta, s_0$) proceeds just like RMAX($N, k, R_{\max}, T, \epsilon, \delta, s_0$), except for the following modifications:

- The algorithm terminates if it discovers a reward greater than $R_{\max}$, more than $k$ actions, or more than $N$ states ($N$, $k$, and $R_{\max}$ can be viewed as the current guesses for these values; if the guess is discovered to be incorrect, the algorithm is restarted with better guesses.)
- if $(s, a_0)$ has just become known, then we set the reward for playing $a_0$ in state $s$ to be $-\infty$ (this ensures that $a_0$ is not played any more in state $s$).

For future reference, we say that *an inconsistency is discovered* if the algorithm terminates because it discovers a reward greater than $R_{\max}$, more than $k$ actions, or more than $N$ states.

**Lemma 5.1:** *Let $K_0$ be defined as above. If the DM plays $a_0$ $K_0$ times at state $s$, then with probability $\geq 1 - \delta/4N$ a new action will be discovered if there is at least one new action at state $s$ to be discovered.*

In the full paper, we show that URMAX($S_0, g_0, D, N, k, R_{\max}, T, \epsilon, \delta, s_0$) is correct provided that the parameters are correct. We get URMAX($S_0, g_0, D, \epsilon, \delta, s_0$) by running URMAX($S_0, g_0, D, N, k, R_{\max}, T, \epsilon, \delta, s_0$) using larger and larger values for $N$, $k$, $R_{\max}$, and $T$. Sooner or later the right values are reached. Once that happens, with high probability, the policy produced will be optimal in all later iterations. However, since we do not know when that happens, we need to continue running the algorithm. We must thus play the optimal policy computed at each iteration enough times to ensure that, if we have estimated $N$, $k$, $R_{\max}$, and $T$ correctly, the average reward stays within $2\epsilon$ of optimal while we are testing higher values of these parameters. For example, suppose that the actual values of these parameters are all 100. Thus, with high probability, the policy computed with these values will give an expected payoff that is within $2\epsilon$ of optimal. Nevertheless, the algorithm will set these parameters to 101 and recompute the optimal policy. While this recomputation is going on, it may get low reward (although, eventually it will get close to optimal reward). We need to ensure that this period of low rewards does not affect the average.

URMAX($S_0, g_0, D, \epsilon, \delta, s_0$):
Set $N := |S_0|, k := |A_0|, R_{\max} := 1, T := 1$
**Repeat forever**
 Run URMAX(($S_0, g_0, D, N, k, R_{\max}, T, \epsilon, \delta, s_0$)
 **if** no inconsistency is discovered
  **then** run the policy computed by
   URMAX(($S_0, g_0, D, N, k, R_{\max}, T, \epsilon, \delta, s_0$) for
   $K_2 + K_3$ steps, where
    where $K_2 = 2(Nk \max(K_1(T+1), K_0))^{\frac{3}{2}} R_{\max}/\epsilon$
    $K_3 = (2R_{\max}+1)\max((\frac{2R_{\max}}{\epsilon})^3, 8\ln(\frac{4}{\delta})^3)/\epsilon$
 $N := N+1; k := k+1, R_{\max} := R_{\max}+1, T := T+1$.

The following theorem shows that URMAX($S_0, g_0, D, \epsilon, \delta, s_0$) is correct. (The proof, which is deferred to the full paper, explains the choice of $K_2$ and $K_3$.)

**Theorem 5.2:** *For all MDPs $M = (S, A, g, P, R)$ compatible with $S_0$ and $g_0$, if the $\epsilon$-return mixing time of $M$ is $T_M$, then for all states $s_0 \in S_0$, with probability at least $1 - \delta$, for all states $s_0 \in S_0$, URMAX($S_0, g_0, D, \epsilon, \delta, s_0$) computes a policy $\pi_{\epsilon, \delta, T_M, s_0}$ such that, for a time $t_{M, \epsilon, \delta}$* that is polynomial in $|S|$, $|A|$, $T_M$, $1/\epsilon$, and $K_0$, and all $t \geq t_{M,\epsilon,\delta}$, we have $U_M(s_0, \pi, t) \geq \text{Opt}(M, \epsilon, T_M) - 2\epsilon$.

Thus, if $\sum_{t=1}^{\infty} D(1,t) = \infty$, the DM can learn to play near-optimally. We now get running time estimates that essentially match the lower bounds of Theorem 4.4.

**Proposition 5.3:** *If $\sum_{t=1}^{T} D(1,t) \geq f(T)$, where $f : [1, \infty) \to \mathbb{R}$ is an increasing function whose co-domain includes $(0, \infty]$, then $K_0 \leq f^{-1}(\ln(4N/\delta))$, and the running time of URMAX is polynomial in $f^{-1}(\ln(4N/\delta))$.*

**Corollary 5.4:** *If $\sum_{t=1}^{T} D(1,t) \geq m_1 \ln(T) + m_2$ (resp., $\sum_{t=1}^{T} D(1,t) \geq m_1 \ln(\ln(T) + 1) + m_2$) for some constants $m_1 > 0$ and $m_2$, then the DM can learn to play near-optimally in polynomial time (resp., exponential time).*

# 6 CONCLUSION

We have defined an extension of MDPs that we call MDPUs, Markov Decision Processes with Unawareness, to deal with the possibility that a DM may not be aware of all possible actions. We provided a complete characterization of when a DM can learn to play near-optimally in an MDPU, and have provided an algorithm that learns to play near-optimally when it is possible to do so, as efficiently as possible. Our methods and results thus provide guiding principles for designing complex systems.

We believe that MDPUs should be widely applicable. We hope to apply the insights we have gained from this theoretical analysis to using MDPUs in practice, for example, to enable a robotic car to learn new driving skills. Our results show that there will be situations when an agent cannot hope to learn to play near-optimally. In that case, an obvious question to ask is what the agent should do. Work on budgeted learning has been done in the MDP setting [6, 7, 12]; we would like to extend this to MDPUs.

**Acknowledgments:** The work of Halpern and Rong was supported in part by NSF grants IIS-0534064, IIS-0812045, and IIS-0911036, and by AFOSR grants FA9550-08-1-0438 and FA9550-09-1-0266, and ARO grant W911NF-09-1-0281.